\documentclass[lettersize,journal]{IEEEtran}
\pdfoutput=1

\usepackage{amsmath,amsfonts,amssymb}
\usepackage{algorithmic}
\usepackage{algorithm}
\usepackage{array}
\usepackage[caption=false,labelfont=sf,textfont=sf]{subfig}
\usepackage{textcomp}
\usepackage{stfloats}
\usepackage{url}
\usepackage{verbatim}
\usepackage{graphicx}
\usepackage{booktabs}

\usepackage{multirow}
\usepackage{bbding}
\usepackage{makecell}
\usepackage{colortbl}
\usepackage{bbm}
\usepackage{setspace}

\usepackage[breaklinks=true,letterpaper=true,colorlinks,linkcolor=blue,citecolor=blue,bookmarks=false]{hyperref}  

\usepackage[numbers, compress]{natbib}
\usepackage{orcidlink}

\definecolor{commentcolor}{RGB}{110,154,155}   

\usepackage[capitalize]{cleveref}
\crefname{section}{Sec.}{Secs.}
\Crefname{section}{Section}{Sections}
\Crefname{table}{Table}{Tables}
\crefname{table}{Tab.}{Tabs.}

\def\eg{\emph{e.g.}} 
\def\ie{\emph{i.e.}} 
 
\def\etc{\emph{etc.}} \def\vs{\emph{vs.}}

\def\etal{\emph{et al.}}

\def\mmethod{ESOD}
\def\moduleA{ObjSeeker}
\def\moduleB{AdaSlicer}
\def\moduleC{SparseHead}
\newcommand{\topa}[1]{\textbf{#1}}

\newcommand{\mdetla}[2][gray]{\scriptsize ~(\textcolor{#1}{#2})}
\newcommand{\rowhl}{\rowcolor[RGB]{230,230,230}}

\newcommand{\editf}[1]{#1}

\begin{document}

\title{ESOD: Efficient Small Object Detection on High-Resolution Images}

\author{
Kai Liu, 
Zhihang Fu, 
Sheng Jin, 
Ze Chen, 
Fan Zhou, 
Rongxin Jiang, 
\\ 
Yaowu Chen, 
and Jieping Ye, 
\IEEEmembership{Fellow,~IEEE}
\thanks{Received 10 October 2023; revised 17 July 2024; accepted 6 November 2024. This work was supported in part by the Fundamental Research Funds for the Central Universities, in part by the Alibaba Cloud through the Research Intern Program, and in part by Zhejiang Provincial Natural Science Foundation of China under Grant LDT23F01013F01. The associate editor coordinating the review of this article and approving it for publication was Dr. Fabrizio Guerrini. (\textit{Corresponding authors: Rongxin Jiang; Jieping Ye}.)}
\thanks{Kai Liu is with the College of Biomedical Engineering and Instrument Science, Zhejiang University, Hangzhou 310027, China. Work was done during his internship at Alibaba Cloud. (E-mail: kail@zju.edu.cn)}
\thanks{Zhihang Fu, Sheng Jin, Ze Chen, and Jieping Ye are with Alibaba Cloud, Hangzhou 310030, China. (E-mail: zhihang.fzh@alibaba-inc.com, jsh.hit.doc@gmail.com, cz265162@alibaba-inc.com, yejieping.ye@alibaba-inc.com)}
\thanks{Fan Zhou and Rongxin Jiang are with the College of Biomedical Engineering and Instrument Science, Zhejiang University, Hangzhou 310027, China, and also with the Zhejiang Provincial Key Laboratory for Network Multimedia Technology, Hangzhou 310027, China. (E-mail: fanzhou@mail.bme.zju.edu.cn, rongxinj@zju.edu.cn)}
\thanks{Yaowu Chen is with the College of Biomedical Engineering and Instrument Science, Zhejiang University, Hangzhou 310027, China, and also with the Embedded System Engineering Research Center, Ministry of Education of China, Hangzhou 310027, China. (E-mail: cyw@mail.bme.zju.edu.cn)}
\thanks{Digital Object Identifier 10.1109/TIP.2024.3501853}

}

\markboth{IEEE TRANSACTIONS ON IMAGE PROCESSING}%
{Shell \MakeLowercase{\textit{et al.}}: A Sample Article Using IEEEtran.cls for IEEE Journals}

\IEEEpubid{0000--0000/00\$00.00~\copyright~2021 IEEE}

\maketitle

\begin{abstract}
Enlarging input images is a straightforward and effective approach to promote small object detection.
However, simple image enlargement is significantly expensive on both computations and GPU memory.
In fact, small objects
are usually sparsely distributed and locally clustered. 
Therefore, massive feature extraction computations are wasted on the non-target background area of images.
Recent works have tried to pick out target-containing regions using an extra network and perform conventional object detection, but the newly introduced computation limits their final performance.
In this paper, we propose to reuse the detector's backbone to conduct feature-level object-seeking and patch-slicing, which can avoid redundant feature extraction and reduce the computation cost.
Incorporating with a sparse detection head, we are able to detect small objects on high-resolution inputs (\eg, 1080P or larger) for superior performance.
The resulting \textit{E}fficient \textit{S}mall \textit{O}bject \textit{D}etection (\mmethod) approach is a generic framework, which can be applied to both CNN- and ViT-based detectors to save the computation and GPU memory costs.
Extensive experiments demonstrate the efficacy and efficiency of our method.
In particular, our method consistently surpasses the SOTA detectors by a large margin (\eg, \textbf{8\%} gains on AP) on the representative VisDrone, UAVDT, and TinyPerson datasets.
Code is available at \url{https://github.com/alibaba/esod}.
\end{abstract}


\section{Introduction}
\label{sec:intro}

\IEEEPARstart{W}{ith} recent advances in convolutional neural networks (CNNs)~\cite{wang2021scaled,chen2020reppoints,cai2019cascade} and Vision Transformers (ViTs)~\cite{dosovitskiy2020image,liu2021swin,carion2020end}, general object detection has promising achieves on public benchmarks including MS COCO~\cite{lin2014microsoft,zhang2023dino} and Pascal VOC~\cite{everingham2010pascal,chen2020slv}.
It has become the foundation for widespread applications, such as autonomous driving~\cite{kitti,sun2020scalability} and security monitoring~\cite{virat,wang2020panda}.
However, detecting small objects (\eg, less than $32\times32$ pixels~\cite{lin2014microsoft}) remains a challenge~\cite{long2017accurate,yu2020scale,li2021gsdet}, 
which hinders visual analysis on specialized platforms like unmanned aerial vehicles (UAVs)~\cite{zhu2018vision,du2023adaptive} and panoramic cameras~\cite{virat,fu2019foreground}. 

To fill the performance gap between detecting small and normal-scale objects, researchers have made numerous efforts on data augmentation~\cite{ghiasi2021simple,yu2020scale}, feature aggregation~\cite{fu2018previewer,gong2021effective}, model evolution~\cite{singh2018sniper,liu2021hrdnet}, \etc~
Whereas, the promotion is still limited, since the poor pixels occupied by small objects lack sufficient visual information to highlight the feature representations~\cite{wu2022uiu}.  
A simple but effective solution is to enlarge the input image's resolution to circumvent the size problem of small objects~\cite{unel2019power,koyun2022focus}.
However, simple resolution-increasing will inevitably lead to explosions of \textit{computation} and \textit{GPU memory}, which is not conducive to the fast detection of small objects in the real world.
\IEEEpubidadjcol

\begin{figure}[tbp]
  \centering
   \includegraphics[width=0.95\linewidth]{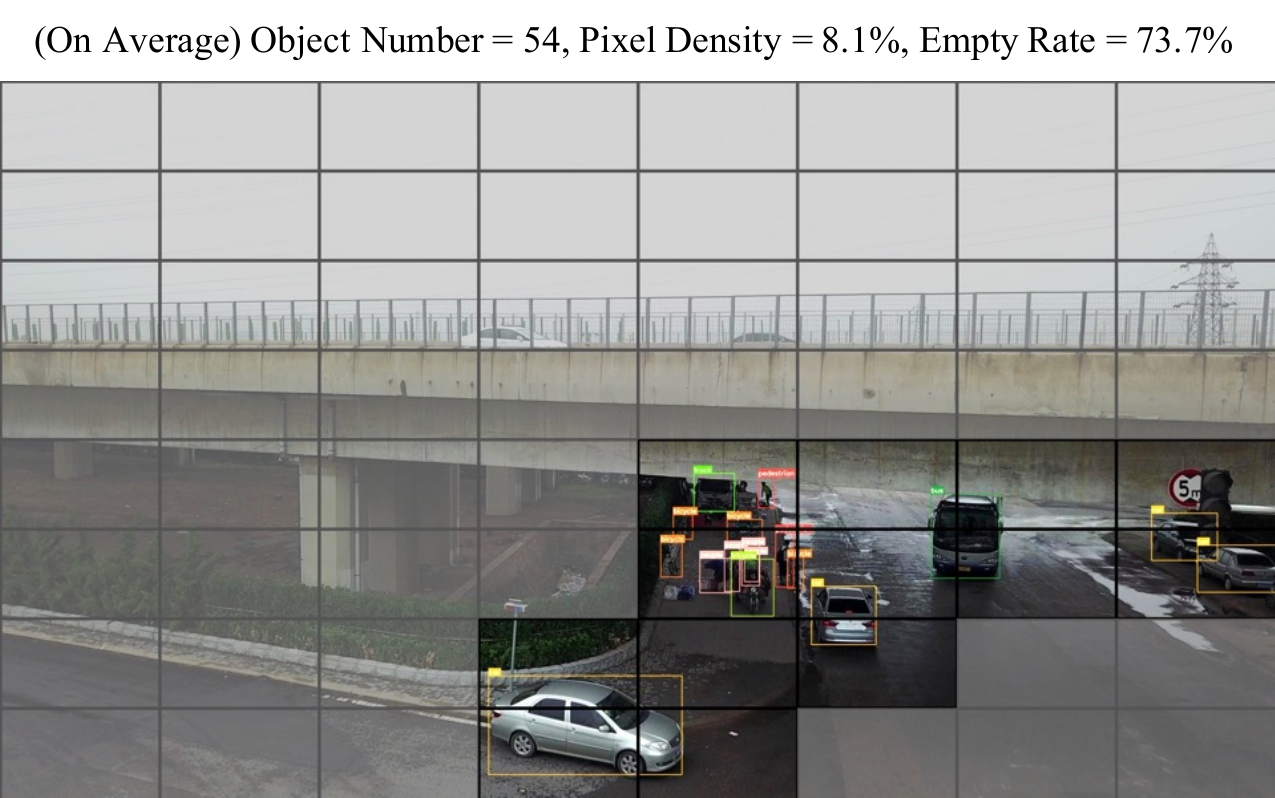}

   \caption{\textbf{Example from VisDrone~\cite{zhu2018vision} dataset.} This image is uniformly sliced into $8~\times~8$ patches. No object exists in most of the patches (masked in gray), while the small objects, \eg, persons, are clustered in one or two patches.}
   \label{fig:intro_moti}
\end{figure}

In fact, most of the computations brought by increased resolution are spent on background regions~\cite{yang2019clustered,koyun2022focus}.
This kind of redundant computation is especially common in practice.
Taking VisDrone~\cite{zhu2018vision} dataset as an example, the targets are sparsely distributed on images captured by UAVs, while small objects tend to be concentrated in specific regions, as shown in \cref{fig:intro_moti}.
On average each image contains 54 targets, but those targets only occupy 8.1\% pixels in total.
We divide each image into $8\times8$ patches uniformly to further explore the target density distribution.
The statistic suggests that over 70\% of the patches contain no objects.
Therefore, it should be more computation friendly to filter out the empty regions at first, rather than process the whole large image without distinction.

\begin{figure}[tbp]
  \centering
   \includegraphics[width=0.85\linewidth]{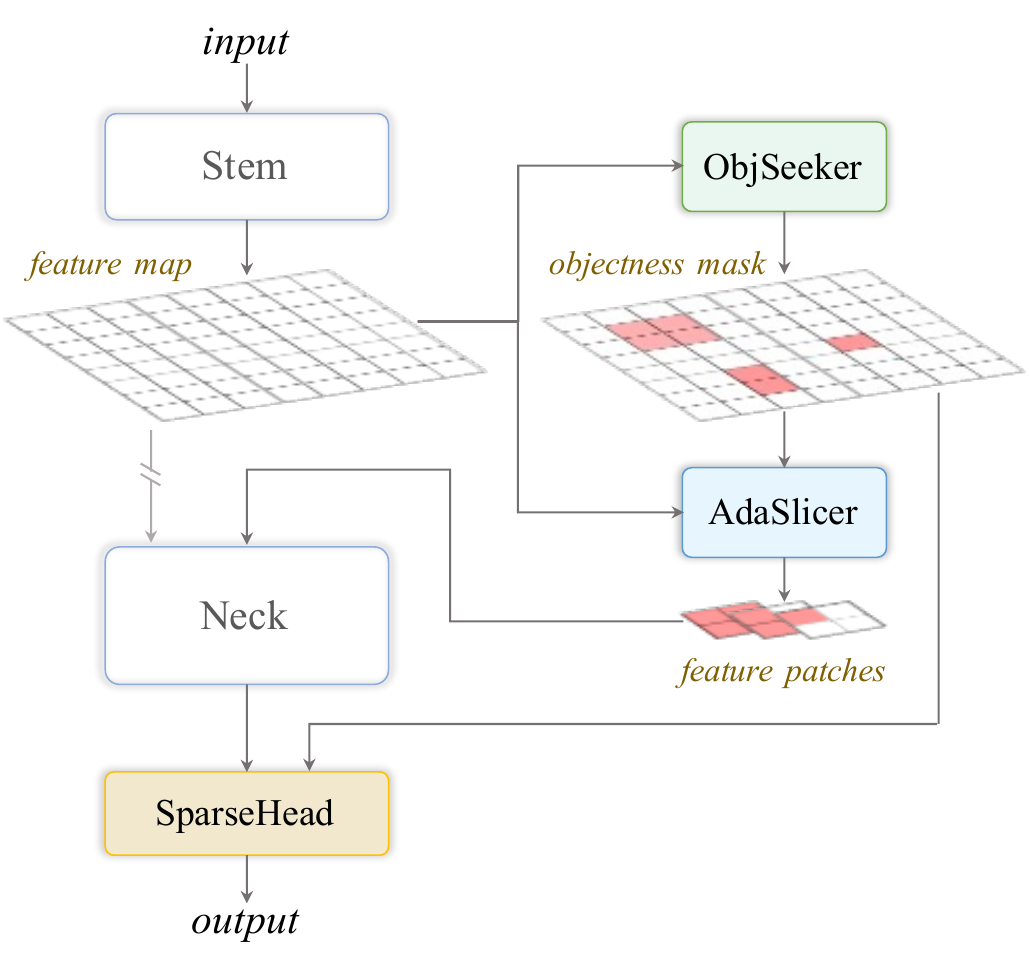}
   \caption{\textbf{Architecture of our generic \mmethod~detector}. \moduleA~is inserted after \textit{stem} to seek a few regions possibly containing objects (colored grids). \moduleB~then slices the feature map into small patches, and discards the background regions. \moduleC~applies sparse detection on the patches.}
   \label{fig:intro_arch}
\end{figure}

In order to efficiently eliminate the backgrounds,
this paper first inserts an \moduleA~module at the early phase of a base detector, to seek the regions that possibly contain objects of interest.
Then, \moduleB~adaptively slices the feature map into small patches with fixed sizes, discards the numerous but non-target feature patches, and feeds the remaining into the following modules for final object prediction. 
Here, \moduleC~applies sparse convolutions~\cite{graham2017submanifold,yan2018second} to the remaining patches to further reduce the computation wasted on backgrounds.
The resulting generic framework \mmethod, as shown in \cref{fig:intro_arch}, boosts modern neural networks in efficiently detecting small objects on high-resolution images.

This idea is motivated by the recent Grounded-SAM\footnote{https://github.com/IDEA-Research/Grounded-Segment-Anything,}
which adopts the ``divide and conquer'' way to segment salient objects with Segment Anything Model (SAM)~\cite{kirillov2023segment} and then endow semantic labels by Grounding-DINO~\cite{liu2023grounding}.
Our proposed \mmethod~coarsely seeks the class-agnostic objects and then determines their category labels and finer localizations.
By doing so, the tremendous computation cost of background regions in high-resolution images is significantly saved.

Indeed, such a filter-then-detect paradigm has recently been studied by researchers~\cite{yang2019clustered,duan2021coarse,cornernetlite}, who usually adopted extra independent networks to generate objectness masks, and sliced the original image into small patches for conventional detectors.
However, the newly-introduced computation is non-negligible due to the redundant feature extraction on original images and sliced image patches.
Hence their preliminary object-seeking is applied to down-sampled images to save computation, whereas small objects with poorer pixels are further filtered. 
By contrast, our \mmethod~conducts object-seeking and image-slicing at the feature level to avoid redundant feature extractions.
In this way, we are able to detect small objects in original high-resolution (\eg, 1080P) or even larger images while maintaining computation and time efficiency.
Therefore, the scarce pixel information of small objects is preserved to the utmost extent.
As a result, our \mmethod~achieves a new state-of-the-art performance on three representative datasets including VisDrone~\cite{zhu2018vision}, UAVDT~\cite{du2018unmanned}, and TinyPerson~\cite{yu2020scale}, consistently improving the efficacy and efficiency. 

It is worth noting that \mmethod~is a plug-and-play optimization approach to both CNNs~\cite{he2016deep,li2022yolov6} and Vision Transformers~\cite{dosovitskiy2020image,carion2020end}. 
The \moduleB~can be easily extended to generate attention masks to avoid quadratically increased computation on visual tokens of background regions.
For more details and experiments please refer to \cref{sec:method_adaslicer} and \cref{sec:exp_res}.

Our contribution can be summarized as follows:

\begin{enumerate} 
    \setlength{\itemsep}{0pt}
    \item We statistically state the fact of sparsely clustered small objects in practice, and conduct feature-level object-seeking with adaptive patch-slicing to avoid redundant feature extraction and save the computation cost. A sparse detection head is adopted to reuse the estimated objectness mask for further computation-saving.
    \item We propose a generic framework \mmethod~that adapts to both CNN and ViT architectures, saving detection computation and GPU memory on high-resolution images.
    \item We surpass state-of-the-art detectors by a large margin (\eg, \textbf{8\%} gains on AP) with comparable computation cost and inference speed on the representative VisDrone~\cite{zhu2018vision}, UAVDT~\cite{du2018unmanned}, and TinyPerson~\cite{yu2020scale} datasets.
\end{enumerate}

\section{Related Works}
\label{sec:related_work}

\subsection{Small Object Detection}
\label{sec:rw_sod}

Inspired by the success of general object detection, many works adopt the ``divide and rule'' idea to address the issue of size variation in small object detection (SOD). SNIP~\cite{singh2018analysis} builds an image pyramid, and at each scale only objects with proper medium size are treated as ground truth. SNIPER~\cite{singh2018sniper} crops several patches with a set of fixed sizes from the original image, avoiding explicitly constructing an image pyramid for multi-scale training. Whereas time-consuming multi-scale testing is required. HRDNet~\cite{liu2021hrdnet} uses a backbone pyramid to utilize the image pyramid, where the heavyweight backbone processes the small image and vice versa, and then takes a multi-scale FPN to fuse extracted features.

In addition, data augmentation is an effective approach to improving the performance of SOD. Simple copy-paste~\cite{ghiasi2021simple} is a strong data augmentation method for instance segmentation and object detection to address various imbalance problems. Yu \etal~propose SM~\cite{yu2020scale} and SM+~\cite{jiang2021sm+} as pre-training strategies to improve the effectiveness of transfer learning by aligning the object size distributions between large source datasets, \eg, MS COCO~\cite{lin2014microsoft}, and small destination dataset. 
Stitcher~\cite{chen2020stitcher} shrinks regular images to generate small objects during training manually, and selectively feeds the collected images to optimize detectors on more small objects.

In contrast, input image enlarging~\cite{unel2019power,wang2020deep} is more effective but lacks efficiency, and our work focuses on saving computation and speeding up detection when enlarging images.  

\subsection{Filter-Then-Detect Paradigm}
\label{sec:rw_ftd}

When taking high-resolution images as input, it is a common practice to filter out several patches from the original image and then perform detection. 
Uniformly slicing the image and enlarging the input patches is a simple but effective way to detect small objects~\cite{unel2019power}. 
To avoid computation on empty patches, ClusDet~\cite{yang2019clustered} employs an extra CPNet to locate the clustered objects and discard the empty regions.
Through estimating a density map independently, DMNet~\cite{li2020density} utilizes sliding windows and connected component algorithms to generate cluster proposals, while CDMNet~\cite{duan2021coarse} applies morphological closing operation and connected regions. 
Meanwhile, UFPMP-Det~\cite{huang2022ufpmp} uses a coarse detector to generate sub-regions, and merges them into a unified image for multi-proxy detection.
And Focus\&Detect~\cite{koyun2022focus} utilizes the Gaussian mixture model to estimate focal regions.

The mentioned methods introduce individual networks to generate cluster regions, crop original images into patches, and feed them to another network for finer object detection. 
However, there exists massive redundant feature extraction in the two networks, which instead damages the detection efficiency.
By contrast, our method performs patch-seeking at the feature level and unifies it with object-detecting in one network. 
No more redundant computation is needed.

\subsection{Sparse Convolution}
\label{sec:rw_spconv}

Sparse CNN~\cite{graham2017submanifold,yan2018second} has recently emerged as a promising solution to accelerate inference by generating pixel-wise sample masks for convolutions.
Perforated-CNN~\cite{figurnov2016perforatedcnns} generates masks with different deterministic sampling methods.
DynamicConv~\cite{verelst2020dynamic} uses a small gating network to predict pixel masks, and SSNet\cite{xie2020spatially} proposes a stochastic sampling and interpolation network.
In particular, sparse convolutions have been applied to the detection head.
QueryDet~\cite{yang2022querydet} builds a cascade sparse query structure to accelerate tiny object detection on high-resolution feature maps,
and CEASC~\cite{du2023adaptive} adaptively adjusts the mask ratio with global feature captured to balance the efficiency and accuracy.

As the above methods usually adopt Gumble-Softmax~\cite{jang2016categorical} or focal loss~\cite{lin2017focal} to train the sparse masks, extra computation cost is introduced.
Instead, this paper proposes a cost-free approach to generating sparse masks to accelerate the detection.

\section{Method}
\label{sec:method}

This section describes our \mmethod~framework to efficiently detect small objects on high-resolution images.
First, we revisit the generic object detector in \cref{sec:method_revisit}, and break down the neural network into three parts (\ie, \textit{stem}, \textit{neck}, and \textit{head}).
Then, we demonstrate the specialized evolution of each part for small object detection in the following sections.

\begin{figure}[tbp]
  \centering
   \includegraphics[width=0.85\linewidth]{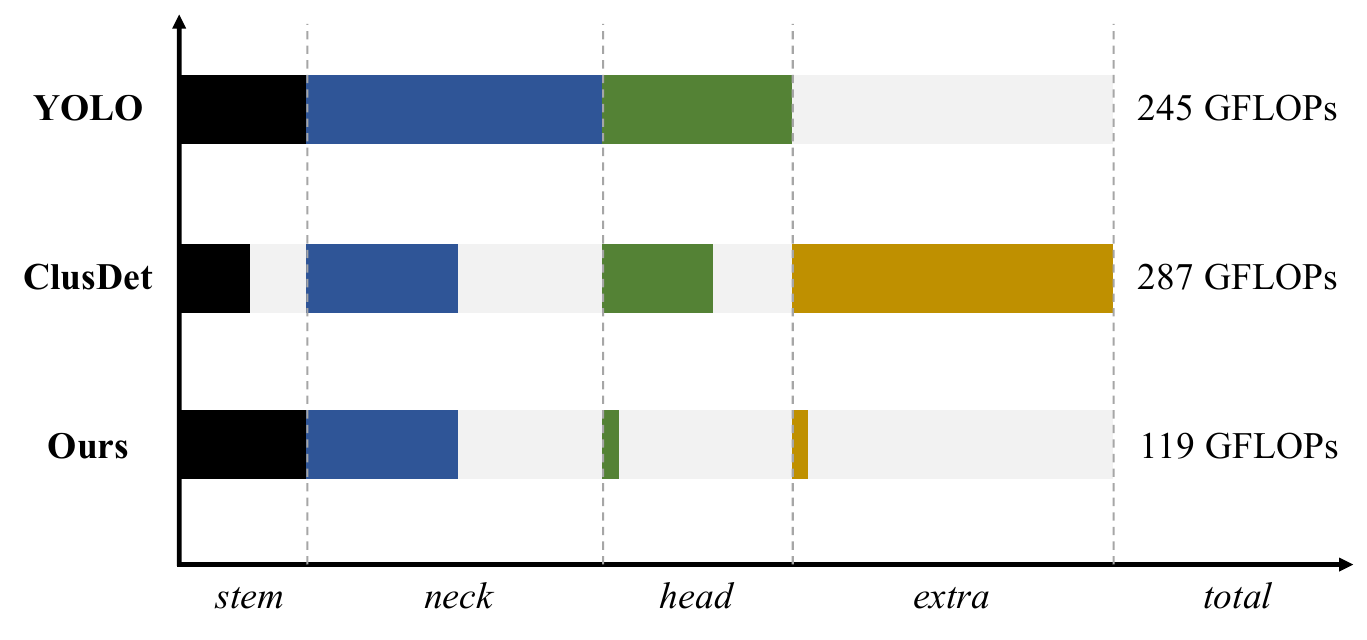}

   \caption{\textbf{Computation distribution when taking inputs at $1,536\times864$.} Though ClusDet~\cite{yang2019clustered} can reduce the computation on the base detector~\cite{glenn_jocher_2021_4679653}, massive extra computation is introduced. Our method avoids this problem.}
   \label{fig:method_cmp}
\end{figure}

\subsection{Revisiting Object Detector}
\label{sec:method_revisit}

In conventional object detectors~\cite{liu2016ssd,zhou2019objects,li2022yolov6}, given an input image $I \in \mathbb{R}^{H \times W \times 3}$, let $\mathbf{P} = \mathcal{F}(I) = \{P_l \in \mathbb{R}^{\frac{H}{2^l} \times \frac{W}{2^l} \times C}\} $ denotes the $l$-level feature pyramid~\cite{lin2017feature} extracted by the backbone $\mathcal{F}$.
Then the detection \textbf{\textit{head}} $\mathcal{H}$ leverages the feature pyramid $\mathbf{P}$ to produce $n$ object predictions: $\mathbf{B} = \mathcal{H}(\mathbf{P}) = \{(x_c^i, y_c^i, w^i, h^i, c^i)\}_{i=1}^n$, where $(x_c, y_c)$ indicates the center coordinates, $(w, h)$ refers to object size, and $c$ denotes category.

Regardless of CNN-based~\cite{liu2016ssd,li2022yolov6} or ViT-based~\cite{carion2020end,zhu2020deformable} networks, the backbone can be divided into two parts (\ie, \textbf{\textit{stem}} and \textbf{\textit{neck}}): $\mathcal{F} \triangleq \mathcal{F}^N \circ \mathcal{F}^S $.
The stem $\mathcal{F}^S$ extracts the preliminary features as $F = \mathcal{F}^S(I)$, and the neck (\eg, FPN-like structures~\cite{lin2017feature,wang2019efficient} or Transformer blocks~\cite{vaswani2017attention,dosovitskiy2020image}) further produces the feature pyramid as $\mathbf{P} = \mathcal{F}^N(F)$.
The overall object detection on image $I$ can be formulated as:

\begin{equation}
\begin{aligned}
  \mathbf{B} = \mathcal{H}(\mathcal{F}^N(\mathcal{F}^S(I)))
  \label{eq:method_det}
\end{aligned}
\end{equation}

\begin{figure*}[tbp]
  \centering
   \includegraphics[width=1.0\linewidth]{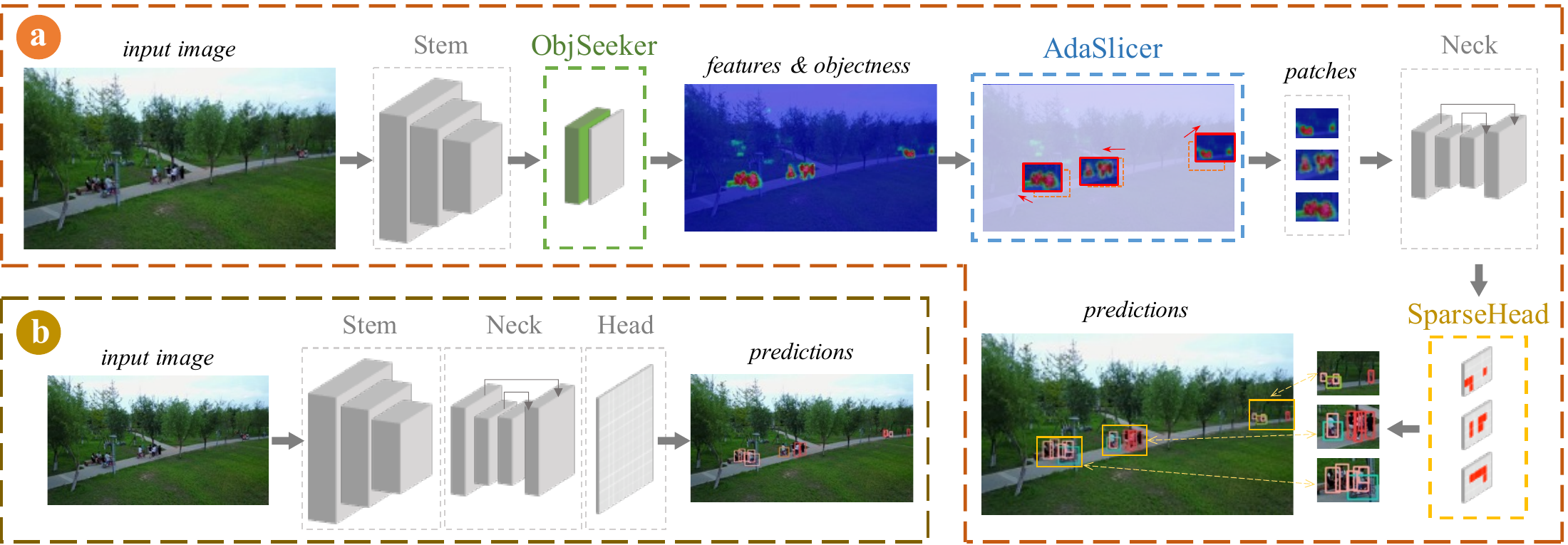}

   \caption{\textbf{Framework comparison between our \mmethod~(a) and conventional detectors (b). }
   For a generic detector, \moduleA~is inserted after \textit{stem} to seek the object-containing regions via estimating a objectness map. \moduleB~then adaptively slices the feature map into small patches, discards the background regions, and sends the remaining to \textit{neck} for feature aggregation. \moduleC~finally applies sparse detection on \textit{head} to save further computation.
   }
   \label{fig:method_fmwk}
\end{figure*}

To speed up the detection process, a common practice~\cite{yang2019clustered,li2020density} is using an extra network to pre-filter the object-containing regions, discard backgrounds, and re-perform object detection on sliced image patches. 
However, as shown in \cref{fig:method_cmp}, the introduced computation by extra networks is unaffordable.
By contrast, this paper reuses the features from the detector itself for efficient object-seeking (\cref{sec:method_objseeker}), adaptively slices the corresponding feature map into small patches (\cref{sec:method_adaslicer}), and leverages sparse detection head on the feature pyramid for further computation-saving (\cref{sec:method_sparsedet}).
The overall framework of our proposed \mmethod~is illustrated in \cref{fig:method_fmwk}.

\subsection{Efficient Object Seeker}
\label{sec:method_objseeker}

The key factor to speed up small object detection is how to efficiently localize the regions that possibly contain objects with less cost. 
Previous works achieve this goal by introducing another independent network to estimate a density map~\cite{li2020density,duan2021coarse} or regress the cluster regions~\cite{yang2019clustered,huang2022ufpmp}. 
However, the redundant computation on feature extraction on preliminary object-seeking and final object-detection is non-negligible. It hinders them from detecting small objects on larger images for better performance. 

To alleviate this problem, we propose to reuse the features from conventional detectors to seek potential objects.
Specifically, an \moduleA~module is inserted after the stem $\mathcal{F}^S$ (\eg, $8\times$ down-sampled) of the detector to estimate the class-agnostic objectness mask $\hat{M}$:

\begin{equation}
\begin{aligned}
  \hat{M} = \mathcal{O}(\mathcal{F}^S(I)) = \mathcal{O}(F) \in \mathbb{R}^{\frac{H}{8} \times \frac{W}{8}}
  \label{eq:method_mask}
\end{aligned}
\end{equation}

\moduleA~$\mathcal{O}$ only comprises a depth-wise separable convolutional (DWConv~\cite{howard2017mobilenets}) block (with BN and ReLU non-linearities), and a standard 1$\times$1 convolutional (Conv) layer.
The kernel size of DWConv is set as 13 to enlarge the receptive field, and the final output channel of Conv is 1.
As shown in \cref{fig:method_cmp}, the introduced \textit{extra} computation (\ie, 1.2 GFLOPs) is negligible compared to the entire detection process.

In fact, an intuitive way to seek potential objects is directly using a region proposal network~\cite{ren2015faster} or cluster proposal network~\cite{yang2019clustered}. 
However, bounding box regression may be not applicable for seeking small objects in network's shallow stage due to the limited feature capacity~\cite{zhang2015cross,ma2019bayesian}. 
Researchers thus leverage the density map to estimate the object distribution~\cite{li2020density,duan2021coarse}.
Whereas, a vanilla density map may damage the preliminary seeking process~\cite{shi2019revisiting}, as our core goal is to seek the foreground (with entire objects) rather than count the crowd (with objects' centers only).

Therefore, our \moduleA~produces the class-agnostic objectness mask to identify foreground from background, instead of generating highly-semantic predictions like cluster coordinates~\cite{yang2019clustered} or object counts (density)~\cite{li2020density}.
This idea is motivated by the recent Grounded-SAM~\cite{liu2023grounding}, which adopts the ``divide and conquer'' way to segment salient objects and then endow semantic labels.
\moduleA~coarsely seeks the class-agnostic objects, and the following modules determine their category labels and finer localizations.

\begin{figure}[b]
  \centering
  \includegraphics[width=1.0\linewidth]{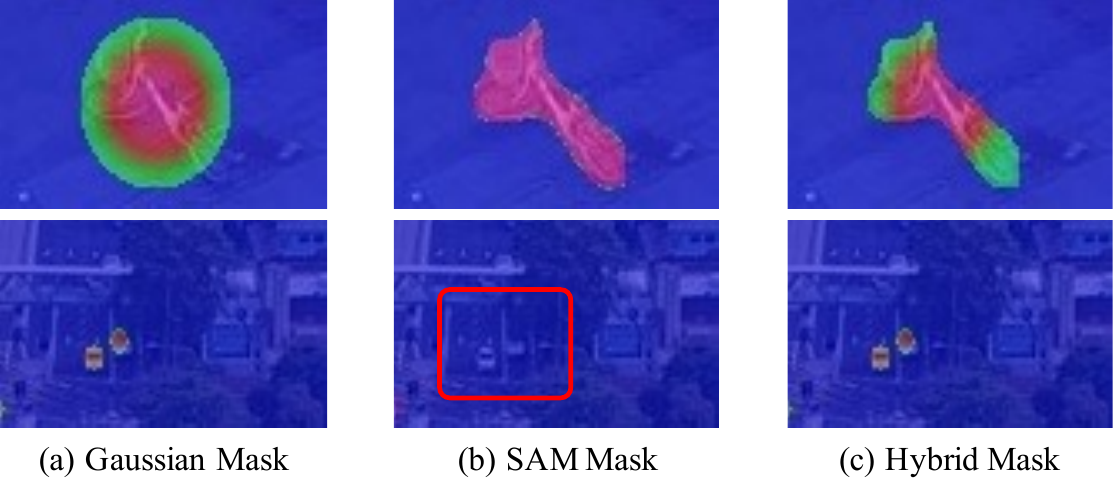}
  \caption{\textbf{Pseudo-labeling strategies to supervise the objectness mask.} Our hybrid strategy (c) utilizes Gaussian masks (a) and SAM~\cite{kirillov2023segment} predictions (b).}
  \label{fig:method_seg_label}
\end{figure}

To learn the \moduleA~module, a hybrid pseudo-labeling strategy is developed to convert bounding box annotations into objectness masking labels.
Given the bounding box ($x_c$, $y_c$, $w$, $h$) only, a commonly-used way is utilizing a Gaussian kernel~\cite{koyun2022focus,yang2022scrdet++} to generate mask labels $M^{G}$:

\begin{equation}
  M^{G}_{x,y} = \exp\left( \frac{1}{2} \left(\frac{(x-x_c)^2}{(w/2)^2} + \frac{ (y-y_c)^2}{(h/2)^2}\right) \times \log{\tau} \right)
  \label{eq:method_mask_gaussian}
\end{equation}

\noindent where the bounding box center becomes $M^{G}_{x_c,y_c} = 1$, and the borders become $M^{G}_{x_c \pm w/2,y_c \pm h/2} = \exp{(\log{\tau})} = \tau$.
Here the threshold $\tau$ is empirically set as 0.5 for foreground-background identification~\cite{kirillov2023segment}.
However, as shown in \cref{fig:method_seg_label}, Gaussian masks fail to capture the exact shape of targets, which motivates us to make an exploratory step to leverage the popular Segment Anything Model (SAM)~\cite{kirillov2023segment}. Specifically, we use SAM to generate high-precision pseudo-masks $M^S = \texttt{SAM}(I) \in [0, 1]^{\frac{H}{8} \times \frac{W}{8}}$ to supplement the \textit{shape prior}.
Despite the extraordinary ability of salient object segmentation, however, SAM still suffers from recognizing small objects, as highlighted in ~\cref{fig:method_seg_label}.
Therefore, this paper proposes a hybrid masking strategy to generate pseudo-labels:

\begin{equation}
  M = \left\{ 
  \begin{aligned}
      M^S \odot M^G,& \quad \textit{if} \; {\Vert M^S \Vert}_1 \; > 0 \; ; \\
      M^G,& \quad \textit{otherwise}
  \end{aligned}
  \right.
  \label{eq:method_mask_hybrid}
\end{equation}

\noindent where $\odot$ refers to Hadamard product.
Following SAM~\cite{kirillov2023segment}, focal loss~\cite{lin2017focal} and dice loss~\cite{milletari2016v} are adopted to optimize the \moduleA~to minimize the discrepancy between \cref{eq:method_mask} and \cref{eq:method_mask_hybrid}. The loss ratio is set to 20:1 as default~\cite{kirillov2023segment}.

\subsection{Adaptive Feature Slicer}
\label{sec:method_adaslicer}

After obtaining the class-agnostic objectness mask $\hat{M}$ from \moduleA, \moduleB~adaptively slices the preliminary feature map $F$ into small patches according to the mask $\hat{M}$, and discards the massive background patches containing no objects.
To achieve this goal, a vanilla solution~\cite{unel2019power} is uniformly slicing the feature $F$ into $k \times k$ patches, and discarding in-activated patches according to objectness mask $\hat{M}$.

\begin{figure}[hb]
  \centering
   \includegraphics[width=1.0\linewidth]{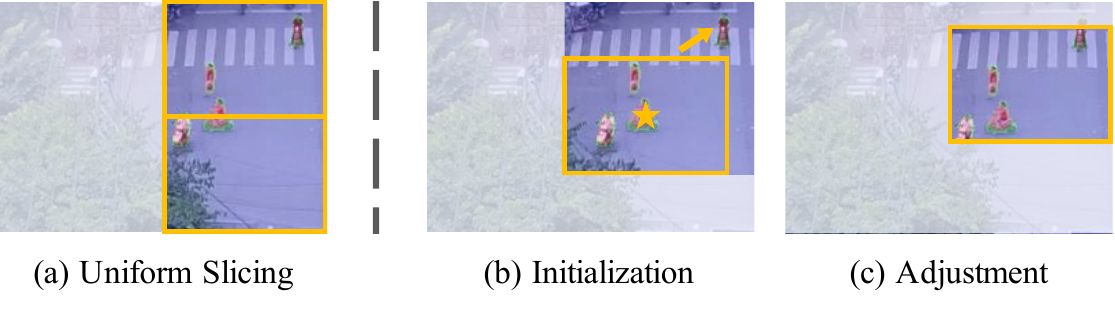}

   \caption{\textbf{Comparison on slicing strategies.} Uniform slicing results in object truncation (a). Our method first initializes the patch at the object center (b) to reduce truncation, and then adjusts it to enclose more objects (c) for efficiency.
   }
   \label{fig:method_slice}
\end{figure}

However, as shown in \cref{fig:method_slice}, such a vanilla slicing strategy 
has two main disadvantages.
First, objects are prone to be cut off into different patches. Though network's receptive field can exceed the feature patches to detect complete objects, large objects are still conducive to being truncated.
Second, such a slicing strategy is actually inefficient, as a large proportion of background still exists in the sliced patches.

\begin{algorithm}[t]
\caption{Adaptive Feature Slicing.}
\begin{algorithmic}
\STATE \textcolor{gray}{Input: $M$, objectness mask}
\STATE \textcolor{gray}{Input: $P = \{W_P, H_P\}$, fixed patch size}
\STATE \textcolor{gray}{Output: $D$, list of patch coordinates}
\STATE 
\STATE $ A \gets M \geq 0.5$
\STATE $ C \gets local\_maxima(M, A) $
\STATE $ S \gets size\_estimate(M, C) $
\STATE $ D \gets \emptyset $
\STATE \textbf{while} $ S \neq \emptyset $ \textbf{do}
\STATE \hspace{.5cm} $ i \gets argmax(S) $
\STATE \hspace{.5cm} $ c \triangleq \{x_c, y_c\} \gets C[i] $
\STATE \hspace{.5cm} $ d \triangleq \{x_1, y_1, x_2, y_2\} \gets calc\_tlbr(c, P) \quad $  \textcolor{gray}{// initiate}
\STATE \hspace{.5cm} $ \delta \triangleq \{\delta_x, \delta_y\} \gets calc\_delta(A, d) $
\STATE \hspace{.5cm} $ d \gets apply\_delta(d, \delta) \qquad\qquad\qquad\quad\;\, $   \textcolor{gray}{// adjust}
\STATE \hspace{.5cm} $ D \gets D \cup \{d\} $
\STATE \hspace{.5cm} $ C, S \gets remove\_covered(C, S, d) \qquad\;\;\, $         \textcolor{gray}{// update}
\STATE \textbf{done}
\end{algorithmic}
\label{alg:method_slice}
\end{algorithm}

\editf{In fact, enclosing all possible objects with the minimum number of sliced patches is an NP-hard problem~\cite{garey1979computers}. We first introduce a greedy strategy in \cref{alg:method_slice}, and then present a simplified alternative for acceleration in \cref{alg:method_slice_sim}}.

As described in \cref{alg:method_slice} and \cref{fig:method_slice} (b), we adaptively slice the feature map $F$ in two steps: initialize a patch box centered at the largest object, and then adjust the patch box to cover as many objects as possible. 
Given the predicted objectness mask $\hat{M}$, one may efficiently locate the object centers \editf{(local maxima)} $\mathbf{C} = \{(x_c^i, y_c^i)\}_{i=0}^{n}$ using a $3\times3$ MaxPooling operation~\cite{zhou2019objects}, and coarsely estimate the object sizes $\mathbf{S} = \{s^i\}_{i=0}^{n}$ using a $9\times9$ AvgPooling operation \editf{(by counting the activated pixels)}. 
During iteration, a patch box is first initialized with a fixed size $(W_p, H_p)$ (equaling to $(\frac{W}{8 \times k}, \frac{H}{8 \times k})$), centered at $(x_c^i, y_c^i)$ with the largest size $s_i$.
The patch's top-left and bottom-right coordinates $(x_1^i, y_1^i, x_2^i, y_2^i)$ become $(x_c^i - W_p/2, y_c^i - H_p/2, x_c^i + W_p/2, y_c^i + H_p/2)$.
Secondly, count the activated locations $\mathbf{A} = \{(x_a^i, y_a^i)\}_{i=1}^{n^{\prime}}$ (indicating the existence of objects) within the patch box.
Then remove the empty regions in the initialized patch box by adjusting the top-left corner $(x_1^i, y_1^i)$ with an offset $(\Delta{x}, \Delta{y})$, where $\Delta{x} = \min\{x_a^i\} - x_1^i$ and $\Delta{y} = \min\{y_a^i\} - y_1^i$.
Finally, remove objects covered by the current patch box $d_i$ from $\mathbf{C}$ and $\mathbf{S}$, and loop until no objects are left (\ie, $\mathbf{C}, \mathbf{S} = \emptyset$).

\begin{algorithm}[ht]
\caption{Simplified Adaptive Feature Slicing.}
\begin{algorithmic}
\STATE 
\STATE \textcolor{gray}{Input: $M$, objectness mask}
\STATE \textcolor{gray}{Input: $G$, patch coordinate candidates}
\STATE \textcolor{gray}{Input: $P = \{W_P, H_P\}$), fixed patch size}
\STATE \textcolor{gray}{Output: $D$, list of patch coordinates}
\STATE 
\STATE $ A \gets M \geq 0.5$
\STATE $ C \gets local\_maxima(M, A) $
\STATE $ D \gets filter\_candi(G, C) \qquad\qquad\qquad\quad\;\;\; $   \textcolor{gray}{// initiate}
\STATE $ \Delta \gets calc\_delta(A, D) $
\STATE $ D \gets apply\_delta(D, \Delta) \qquad\qquad\qquad\qquad $   \textcolor{gray}{// adjust}
\STATE $ D \gets remove\_overlap(D) $
\end{algorithmic}
\label{alg:method_slice_sim}
\end{algorithm}

\editf{However, the slicing strategy introduced in \cref{alg:method_slice} selects foregrounds iteratively, which may harm the overall inference latency without GPU accelerating. 
Therefore, we present a simplified alternative in \cref{alg:method_slice_sim} to slice patches in a parallel way.
It also adaptively slices the feature maps in an initiate-then-adjust manner, but the patch candidates are initialized from pre-processed uniform slicing and adjusted in a single round.
Specifically, after finding the activation regions $\mathbf{A}$ and potential object centers $\mathbf{C} = \{(x_c^i, y_c^i)\}_{i=0}^{n}$, we preserve the patch boxes that contain at least one object center $\{(x_c^i, y_c^i)\}$ as patch candidates and discard the others.
In this case, \cref{fig:method_slice} (a) serves as the initialization step in \cref{alg:method_slice_sim}.
Then, we calculate and apply the offsets (as described before) for each patch candidate, and remove overlapped patches (\eg, the two patch boxes in \cref{fig:method_slice} (a) can be adjusted to the same position in \cref{fig:method_slice} (c)) to avoid redundant computation.
Despite the sub-optimal performance (\eg, possible truncation in large objects), the simplified strategy removes the loop and thus can be paralleled by GPU for further acceleration.
}

With either \cref{alg:method_slice} or \cref{alg:method_slice_sim}, our \moduleB~$\mathcal{A}$ regularizes the feature map $F$ as $N_P$ feature patches:

\begin{equation}
  F_P = \mathcal{A}(F, \hat{M}) \in \mathbb{R}^{N_P \times H_P \times W_P}
  \label{eq:method_feat_slice}
\end{equation}

As shown in \cref{fig:method_fmwk}, only the feature patches containing objects are passed into the following neck of the detector for further feature aggregation.
The background regions are greatly discarded, and meaningless computation is saved.

\begin{figure}[H]
  \centering
   \includegraphics[width=1.0\linewidth]{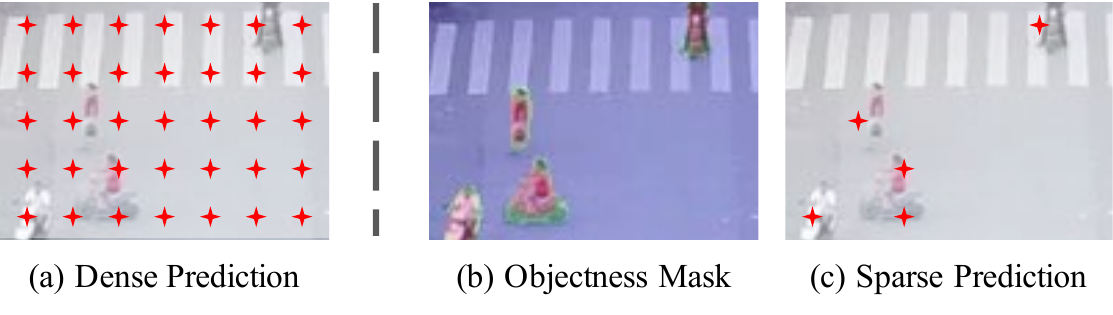}

   \caption{\textbf{Illustration of sparse detection.} Compared to the dense prediction (a), the predicted objectness mask (b) provides potential object centers for \moduleC~to apply sparse detection (c) to save computation. }
   \label{fig:method_sparse}
\end{figure}

In ViT-based detectors~\cite{carion2020end}, the patch size $W_p \times W_p$ becomes $1\times1$, where the feature patch becomes a single image token. 
And the above \cref{alg:method_slice} is not even needed.
To save the massive computations wasted on background regions in Transformer blocks (especially the self-attention module), one may simply preserve the activated tokens in $\mathbf{A} = \{(x_a^i, y_a^i)\}_{i=1}^{m}$ and discard the remaining.
The computation and GPU memory in the neck of detectors are significantly reduced.

\subsection{Sparse Detection Head}
\label{sec:method_sparsedet}

In conventional object detectors, the decoupled detection head on aggregated high-resolution feature maps is proved essential for detecting small objects~\cite{li2022yolov6,ge2021yolox}. 
However, the computation cost remains a problem to address, especially on resource-constrained platforms.

Recently, sparse convolutions~\cite{graham2017submanifold,yan2018second} show a promising solution by only operating convolutions on sparsely sampled grids in the feature maps.
However, previous works~\cite{yang2022querydet,du2023adaptive} obtain the sparse locations via learnable masks, where extra parameters and computation are introduced
\editf{and the optimization difficulty increases accordingly}.

On the contrary, our \moduleC~$\mathcal{H}^{sp}$ directly applies sparse convolutions on the possible object centers $\mathbf{C} = \{(x_c^i, y_c^i)\}_{i=0}^{n}$ (obtained from the objections mask $\hat{M}$ estimated by \moduleA) in the aggregated feature patches for object detection:

\begin{equation}
  B^{\prime} = \mathcal{H}^{sp}(\mathcal{F}^N(F_P))
  \label{eq:method_sparse_det}
\end{equation}

The process is illustrated in \cref{fig:method_sparse}.
The computation on the detection head is largely saved in a cost-free way.

Combining with \moduleA, \moduleB, and \moduleC, our proposed \mmethod~framework is able to efficiently detect small objects on high-resolution inputs for better performance.
\editf{\moduleA~is optimized with the detector network together, while \moduleB~and \moduleC~are training-free components as they do not introduce new learnable parameters.
In particular, we use the ground-truth objectness mask for feature slicing in preliminary warm-up steps to ensure training stability, and in the remaining steps, we use \moduleA's predicted objectness mask for the following procedures.}

\section{Experiment}
\label{sec:exp}

\subsection{Datasets}
\label{sec:exp_dataset}

To validate the effectiveness of the proposed \mmethod~and reduce the bias, we conduct a series of experiments on three popular and representative datasets: VisDrone~\cite{zhu2018vision}, UAVDT~\cite{du2018unmanned}, and TinyPerson~\cite{yu2020scale}.

\textbf{VisDrone.} The videos and images in VisDrone are captured by multiple drone platforms in diverse scenarios across fourteen cities. The VisDrone-DET dataset contains 10,209 static images (6,471 for training, 548 for validation, and 3,190 for testing) with 10 object categories in total.
The image scale ranges from 960$\times$540 to 2,000$\times$1,500.
When uniformly dividing images into $8\times8$ patches, as shown in \cref{fig:intro_moti}, over 70\% of the patches contain no objects, indicating the sparse distribution.
Following the literature~\cite{yang2019clustered,yang2022querydet,du2023adaptive}, we take the validation set for evaluation.

\textbf{UAVDT.} Acquired by a drone platform at several locations in urban areas, the UAVDT dataset consists of 50 video sequences with 3 categories to detect. 
There are 30 videos (23,258 frames) for training and 20 (15,069 frames) for testing, whose resolution is 1,024$\times$540. 
On average there are 18 objects in one image with only 4.9\% pixels occupied, and around 84\% patches are empty.

\textbf{TinyPerson.} All of the images in TinyPerson are collected from the Internet, which mainly focus on persons around the seaside, where the sizes of objects are both absolutely and relatively small (\eg, less than $20\times20$ pixels). 
The dataset contains 1,610 images (794 for training and 816 for testing) mainly with a size of 1,920$\times$1,080.
On average each image contains 25 objects, which only occupy 0.87\% pixels. These objects are more sparsely distributed in the image, and the patches' empty rate even reaches up to 89\%.

\subsection{Evaluation Protocols}
\label{sec:exp_eval}

To evaluate the performance of detectors, AP (average precision) is taken as the primary metric. AP$_{50}$ refers to the area under the precision-recall curve averaged by all categories with the IoU threshold of 0.5,
and AP is computed by averaging precision under IoU thresholds ranging from 0.5 to 0.95 with a step of 0.05.
AP$^s$ is adapted from COCO~\cite{lin2014microsoft}, which measures the AP on small objects under 32$\times$32 pixels. 
Specifically, AP$_{50}^{t}$ and AP$_{50}^{s}$ for TinyPerson~\cite{yu2020scale} respectively computes the AP$_{50}$ on tiny objects from 2$\times$2 pixels to 20$\times$20 pixels and AP$_{50}$ on small objects from 20$\times$20 pixels to 32$\times$32 pixels. 

To validate the effectiveness of our \mmethod, \editf{we compute the FLOPs (floating-point operations) with the popular 
fvcore library\footnote{https://github.com/facebookresearch/fvcore}.
We report the average FLOPs on each input image in the validation set as a proxy to quantify the computational complexity.}
Besides, we test the speed of \editf{all the detectors (including the compared models with their official codes)} on an Nvidia V100 GPU with a batch size of 1 \editf{for a consistent comparison}, and report the FPS in the following sections.

In particular, as our method concentrates on coarsely seeking the objects for feature-slicing and locating their centers for sparse detection, two specific metrics are developed for the ablation study in \cref{sec:exp_abl}, \ie, Best Possible Recall for objects' bounding boxes (BPR$^{\texttt{box}}$) and Best Possible Recall for objects' centers (BPR$^{\texttt{ctr}}$):

\begin{equation}
\begin{aligned}
  \text{BPR}^{\texttt{box}} &= \frac{1}{N} \sum_{i=1}^{N} \mathbbm{1}\left\{ \frac{\left| box^i \cap patch^j \right|}{\left| box^i \right|} > 0.5 \right\} \\
  \text{BPR}^{\texttt{ctr}} &= \frac{1}{N} \sum_{i=1}^{N} \mathbbm{1}\left\{ (x_c^i, y_c^i) \in \mathbf{C} \right\}
  \label{eq:metric_bpr}
\end{aligned}
\end{equation}

\noindent where BPR$^{\texttt{box}}$ measures the ratio of objects with more than 50\% of the area enclosed by an arbitrary patch, and BPR$^{\texttt{ctr}}$ calculates the ratio of objects whose center is in the local-maxima collection $\mathbf{C}$ on the predicted objectness mask $\hat{M}$.

\subsection{Implementation Details}
\label{sec:exp_impl}

We implement our method based on vanilla PyTorch~\cite{paszke2019pytorch}, and all the models are trained on two Nvidia V100 GPUs.
To construct the baseline, we equip the novel YOLOv5~\cite{glenn_jocher_2021_4679653} with a decoupled detection head~\cite{li2022yolov6} for better performance~\cite{ge2021yolox,li2022yolov6}.
With the proposed \moduleA, \moduleB, and \moduleC, our \mmethod~is developed.
All the detectors are trained for 50 epochs with the default settings (\eg, an SGD optimizer with a weight decay of 0.0005, and an initial learning rate of 0.01 with the cosine annealing schedule).
The batch size is 8.
Unless otherwise specified, we employ the medium-size backbone~\cite{glenn_jocher_2021_4679653} to build the detectors, and the larger side of input size is set to 1,536 for VisDrone, 1,280 for UAVDT, and 2,048 for TinyPerson, respectively.

\subsection{Main Results}
\label{sec:exp_res}

\begin{table}[tb]
    \centering
    \small
    \setlength{\tabcolsep}{5pt}
    \caption{Performance comparison against SOTA detectors on VisDrone and UAVDT datasets. ``$\dagger$'' means simply $1.25\times$ enlarging the input sizes (both of width and height). Except the GFLOPs metric, higher AP, AP$_{50}$, and FPS are better. Best results are marked in \textbf{Bold}.}
    \label{tab:exp_uav}
    \begin{tabular}{c|l|cc|cc}
        \toprule
        Dataset & Detector & AP & AP$_{50}$ & GFLOPs & FPS  \\
        \midrule
        \multirow{9}{*}{VisDrone}   
        & FASF~\cite{zhu2019feature}       & 26.3 & 50.3 & 518.2 & 15.3 \\
        & ClusDet~\cite{yang2019clustered}           & 26.7 & 50.6 & 436.0 & 6.3 \\
        & DMNet~\cite{li2020density}               & 28.2 & 47.6 & 471.4 & 5.9 \\
        & CDMNet~\cite{duan2021coarse}     & 29.2 & 49.5 & $-$   & $-$ \\
        & UFPMP-Det~\cite{huang2022ufpmp}  & 36.6 & \topa{62.4} & 658.7 & 8.5 \\
        & QueryDet~\cite{yang2022querydet} & 28.3 & 48.1 & 888.4 & 8.6 \\
        & CEASC~\cite{du2023adaptive}      & 28.7 & 50.7 & 150.2 & 26.9 \\
        \cmidrule(lr){2-6}
        & \textbf{\mmethod} (Ours) & 36.0 & 59.7 & \topa{119.5} & \topa{36.4} \\
        & $\dagger$ \textbf{\mmethod} (Ours) & \topa{37.9} & 62.3 & 180.6 & 28.6 \\
        \midrule
        \midrule
        \multirow{7}{*}{UAVDT} 
        & FasterRCNN~\cite{ren2015faster} & 11.0 & 23.4 & 249.9 & 27.8 \\
        & ClusDet~\cite{yang2019clustered}          & 13.7 & 26.5 & 421.7 & $-$ \\
        & DMNet~\cite{li2020density}              & 14.7 & 24.6 & 575.3 & $-$ \\
        & CDMNet~\cite{duan2021coarse}    & 16.8 & 29.1 & $-$   & $-$   \\
        & CEASC~\cite{du2023adaptive}     & 17.1 & 30.9 & 64.1 & 35.6 \\
        \cmidrule(lr){2-6}
        & \textbf{\mmethod} (Ours) & 22.5 & 40.7 & \topa{43.7} & \topa{41.1} \\
        & $\dagger$ \textbf{\mmethod} (Ours) & \topa{23.6} & \topa{47.6} & 68.6 & 36.7 \\
        \bottomrule
    \end{tabular}
\end{table}

\begin{table}[tb]
    \centering
    \small
    \setlength{\tabcolsep}{4pt}
    \caption{Performance comparison against SOTA detectors on TinyPerson benchmark.}
    \label{tab:exp_tod}
    \begin{tabular}{c|l|cc|cc}
        \toprule
        Dataset & Detector & AP$^{t}_{50}$ & AP$^{s}_{50}$ & GFLOPs & FPS  \\
        \midrule
        \multirow{7}{*}{TinyPerson} 
        & RetinaNet~\cite{lin2017focal}       & 33.5 & 48.3 & 515.4 & 15.9 \\
        & FasterRCNN~\cite{ren2015faster}     & 47.3 & 63.2 & 492.8 & 16.5 \\
        & ScaleMatch~\cite{yu2020scale}       & 51.3 & 67.0 & 491.1 & 16.9 \\
        & ScaleMatch+~\cite{jiang2021sm+}     & 52.6 & 67.4 & 486.7 & 18.3 \\
        & CascadeRCNN~\cite{cai2019cascade}   & 54.7 & 70.1 & 739.0 & 12.2 \\
        \cmidrule(lr){2-6}
        & \textbf{\mmethod} (Ours)     & 61.3 & 74.4 & \topa{148.3} & \topa{32.8} \\
        & $\dagger$ \textbf{\mmethod} (Ours) & \topa{64.0} & \topa{75.8} & 234.5 & 24.3 \\
        \bottomrule
    \end{tabular}
\end{table}

\textbf{Comparison with state-of-the-art methods.} 
We compare our method with the SOTA rivals on three datasets in \cref{tab:exp_uav} and \cref{tab:exp_tod}. 
Our \mmethod~consistently surpasses the competitors by a large margin on both accuracy and efficiency.

Specifically, compared to detectors who adopt the image-level filter-then-detect paradigm (including ClusDet~\cite{yang2019clustered}, DMNet~\cite{li2020density}, and CDMNet~\cite{duan2021coarse}), our \mmethod~achieves superior performance on both VisDrone~\cite{zhu2018vision} and UAVDT~\cite{du2018unmanned} datasets, as shown in \cref{tab:exp_uav}.
It indicates that our feature-level object-seeking and patch-slicing are more efficient since the massive redundant feature extraction is avoided. 
Thus \mmethod~is able to enlarge the input resolution (\eg, $1.25\times$, denoted as $\dagger$) for better detection performance with affordable computation cost.

\begin{figure}[tb]
  \centering
  \includegraphics[width=1.0\linewidth]{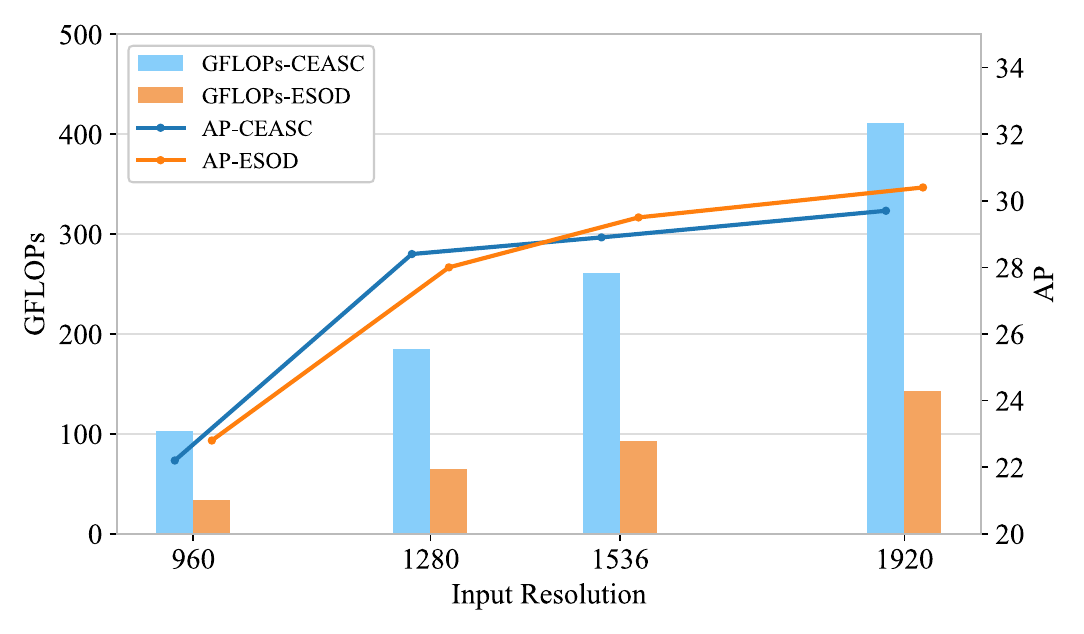}

  \caption{\textbf{Performance comparison to CAESC~\cite{du2023adaptive}. With the same ResNet-18~\cite{he2016deep} backbone, our \mmethod~consistently saves 2/3 of GFLOPs overload as the input resolution extends while maintaining comparable AP results.}}
  \label{fig:res_range}
\end{figure}

Compared with recent SOTA methods like QueryDet~\cite{yang2022querydet} and CEASC~\cite{du2023adaptive} that merely employ sparse detection head for acceleration, our approach saves significant computations on background areas during feature extraction and aggregation via \moduleA~and \moduleB.
\cref{fig:res_range} further illustrates that with the same backbone against CEASC, our method stably reduces the GFLOPs by 2/3 regardless of the varying input resolution.
Thus, \mmethod~is able to enlarge the input image for superior performance (\eg, 47.6 \vs~30.9 of AP$_{50}$ on UAVDT) while maintaining computation costs (\eg, 68.6 \vs~64.1 of GFLOPs) in \cref{tab:exp_uav}.
\editf{We have also noticed that UFPMP-Det~\cite{huang2022ufpmp} surpasses our method by a negligible 0.1 AP$_{50}$, in trade of an unacceptable efficiency degradation from 28.6 FPS to 8.5 FPS. The main reason is that UFPMP-Det employs a separate Faster-RCNN model for image-level coarse region detection and adopts another RetinaNet model for final object detection, where massive redundant computation exists and the two detectors are unable to work in an end-to-end manner, harming the computational complexity and overall latency.}

Besides, \cref{tab:exp_tod} demonstrates heavy networks and sophisticated strategies (like Cascade RCNN~\cite{cai2019cascade}) are unnecessary for small object detection~\cite{yu2020scale}. 
Simple resolution-enlarging is still an effective way for better performance, and our \mmethod~can significantly save the computation and speed up the inference.

\begin{table}[tb]
    \centering
    \small
    \setlength{\tabcolsep}{5pt}
    \caption{\editf{Adaptation to various CNN- and ViT-based baseline detectors on the VisDrone dataset. ``$^*$'' means the model is trained on inputs at 960 while tested at 1,536.}}
    \label{tab:exp_arch}
    \begin{tabular}{c|l|ccc|cc}
        \toprule
        Arch & Detector & AP$^{s}$ & AP & AP$_{50}$ & GFLOPs & FPS  \\
        \midrule
        \multirow{15}{*}{CNN} 
        & RetinaNet~\cite{lin2017focal} & 18.7 & 26.1 & 46.2 & 340.6 & 18.8 \\
        & QueryDet~\cite{yang2022querydet} & - & 26.5 & 46.5 & 321.2 & 19.6 \\
        & \textbf{\mmethod} & 18.5 & 26.0 & 45.9 & \topa{172.9} & \topa{30.0} \\
        & $\dagger$ \textbf{\mmethod} & \topa{20.2} & \topa{27.5} & \topa{48.1} & 278.1 & 23.9 \\
        \cmidrule(lr){2-7}
        & YOLOv5~\cite{glenn_jocher_2021_4679653} & 28.5 & 36.2 & 60.1 & 264.9 & 26.1 \\
        & \textbf{\mmethod} & 28.3 & 36.0 & 59.7 & \topa{119.5} & \topa{36.4} \\
        & $\dagger$ \textbf{\mmethod} & \topa{30.8} & \topa{37.9} & \topa{62.3} & 180.6 & 28.6 \\
        \cmidrule(lr){2-7}
        & RTMDet~\cite{lyu2022rtmdet} & 28.7 & 36.5 & 60.5 & 252.8 & 28.6 \\
        & \textbf{\mmethod} & 28.4 & 36.2 & 60.1 & \topa{110.0} & \topa{37.2} \\
        & $\dagger$ \textbf{\mmethod} & \topa{30.7} & \topa{38.1} & \topa{62.4} & 171.1 & 29.6 \\
        \cmidrule(lr){2-7}
        & YOLOv8~\cite{Jocher_Ultralytics_YOLO_2023} & 29.5 & 37.5 & 61.3 & 323.4 & 24.3 \\
        & \textbf{\mmethod} & 29.4 & 37.3 & 61.0 & \topa{147.0} & \topa{33.3} \\
        & $\dagger$ \textbf{\mmethod} & \topa{31.1} & \topa{38.4} & \topa{62.9} & 232.9 & 25.2 \\
        \midrule
        \midrule
        \multirow{7}{*}{ViT} 
        & $^*$Vanilla~\cite{dosovitskiy2020image} & 23.5 & 30.9 & 53.7 & 842.7 & 3.1 \\
        & \textbf{\mmethod} & 26.0 & 33.9 & 57.9 & \topa{248.5} & \topa{26.1} \\
        & $\dagger$ \textbf{\mmethod} & \topa{29.0} & \topa{35.6} & \topa{60.4} & 406.1 & 20.0 \\
        \cmidrule(lr){2-7}
        & GPViT~\cite{yang2023gpvit} & 30.3 & 37.6 & 62.8 & 1242.4 & 4.8 \\
        & \textbf{\mmethod} & 29.9 & 37.0 & 61.7 & \topa{474.8} & \topa{13.7} \\
        & $\dagger$ \textbf{\mmethod} & \topa{31.3} & \topa{38.3} & \topa{63.0} & 876.0 & 7.1 \\
        \bottomrule
    \end{tabular}
\end{table}

\textbf{Adaptation to various detectors across architectures.} 
To evaluate our method's versatility and universality, we 
\editf{extend our \mmethod~to a variety of popular object detectors across the conventional convolutional neural networks (CNN) to the recent vision transformers (ViT), as verified in \cref{tab:exp_arch}.}

\editf{
For CNN-based object detectors, we start our investigation with the widely-used RetinaNet model~\cite{lin2017focal}.
In particular, we first train the standard RetinaNet with detection heads on P3-P7 layers in \cref{tab:exp_arch}, and simultaneously test QueryDet~\cite{yang2022querydet} at the same input resolution of 1,536 for a fair comparison. 
The results show that although QueryDet boosts 0.4 AP with the newly-introduced detection head and on the high-resolution P2 layer, the overall GFLOPs cost and inference speed remain nearly unchanged because of the extra computation (CEASE~\cite{du2023adaptive} has observed similar phenomenons).
Instead, our \mmethod~can significantly reduce the computation from 340.6 to 172.9 GFLOPs and accelerate to 1.6$\times$, as we introduce no heavy heads and further reduce the useless computation (on background areas) by \moduleA~and \moduleB~in the feature extraction stage.
When simply 1.25$\times$ enlarging the input resolution to 1,920 (denoted as $\dagger$), our \mmethod~outperforms QueryDet by a large margin with fewer computational costs.
}

\editf{
Then, we extend to study more recently advanced models, including YOLOv5~\cite{glenn_jocher_2021_4679653},  RTMDet~\cite{lyu2022rtmdet}, and YOLOv8~\cite{Jocher_Ultralytics_YOLO_2023}.
According to \cref{tab:exp_arch}, our \mmethod~can well adapt to different baseline detectors.
With the same input resolution, \mmethod~significant reduces the computation and speeds up the inference, and as the input resolution enlarges, our \mmethod~achieves a consistent enhancement in detection performance (\eg, 1.5-2.3 gains on AP$^s$) with even higher inference throughput (+1 FPS).
When the baseline model becomes stronger (\eg, from RetinaNet to YOLOv8), our \mmethod~brings consistent improvement in object detection's efficiency and efficacy, indicating our generalizability to incorporate with more advanced detectors in the future literature.
}

\begin{figure}[tb]
  \centering
  \includegraphics[width=1.0\linewidth]{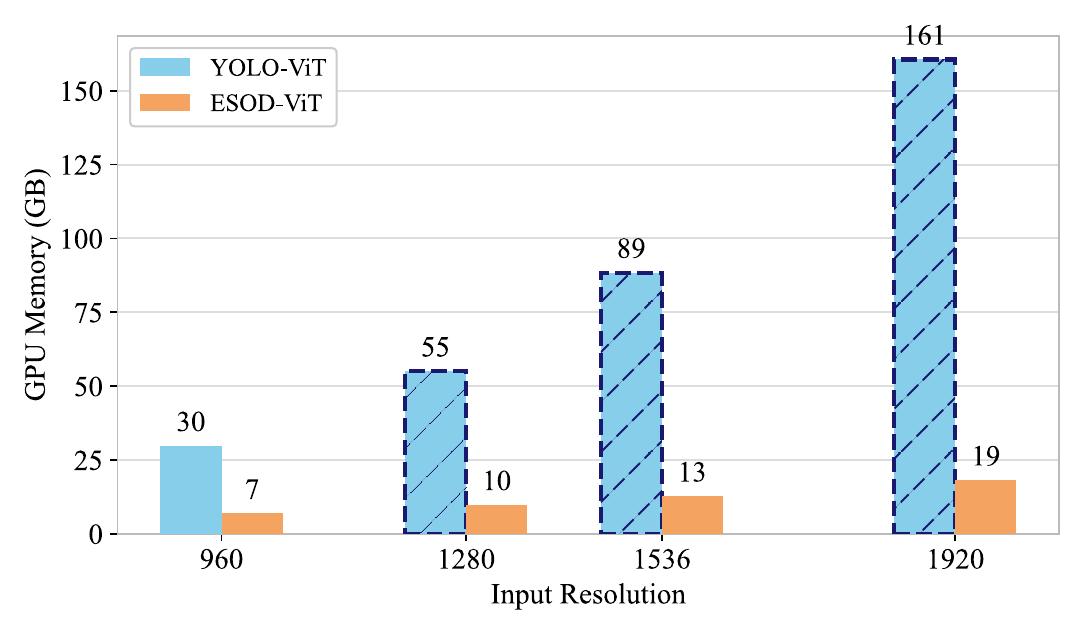}

  \caption{\textbf{GPU memory overhead (GB) in training at batch-size of 1.} Conventional ViT backbone makes the cost generally prohibitive on high-resolution images. Our \mmethod~reduces the cost dramatically from 161 GB to an affordable 19 GB with the assistance of \moduleA~and \moduleB.}
  \label{fig:abl_gpu_mem}
\end{figure}

For ViT-based detectors, we first build a vanilla detector by replacing the CSP Block~\cite{wang2020cspnet} in the YOLO~\cite{glenn_jocher_2021_4679653} baseline with the Transformer Block~\cite{dosovitskiy2020image,vaswani2017attention}, which consists of a Multi-Head Self-Attention (MHSA) layer and a Feed-Forward Network (FFN) layer.
However, despite the promising performance by ViT~\cite{dosovitskiy2020image,liu2021swin}, the computation and GPU memory cost \textit{quadratically} increases as the input size is enlarged (mainly because of the MHSA layer). 
When detecting small objects on high-resolution images, the training cost is generally prohibitive (\eg, 161 GB GPU memory required for an input size of $1,920\times1,920$, as illustrated in \cref{fig:abl_gpu_mem}).
Therefore, the baseline model with ViT architecture can only be trained on $960\times960$, resulting in poor detection performance, as shown in \cref{tab:exp_arch}.

In contrast, \cref{fig:abl_gpu_mem} demonstrates our \mmethod~can significantly reduce the GPU memory cost (\eg, only 19 GB required on $1,920\times1,920$), which enables training and testing the detector on a higher resolution for better performance (\eg, 29.0 \vs~23.5 of AP$^s$). 
In fact, \cref{fig:abl_gpu_mem} implies the GPU memory required by \mmethod~is almost linearly increased as the input size grows, mainly because the object number in images is constant whenever the input resolution changes. 
As described in \cref{sec:method_adaslicer}, \mmethod~only performs self-attention on potential object regions, and the computation and GPU memory merely grow in the preliminary feature extraction process.
Our method thus makes the training and inference on high-resolution images affordable.

\editf{
Furthermore, we adapt \mmethod~to the GPViT~\cite{yang2023gpvit} backbone that perceives more high-resolution information, to investigate further improvements on ViT-based detectors.
The experimental results are displayed in \cref{tab:exp_arch}.
Accordingly, our method can also significantly reduce 60\% computations and speed up the inference for 2.8$\times$ when taking the same input resolution, and a simple input-enlargement operation builds a new state-of-the-art small object detection performance (\eg, 31.3 of AP$^s$).
However, the inference cost becomes gradually unaffordable (\eg, lower than 10 FPS), which highly depends on engineering optimization and hardware resources.
Meanwhile, at the same input size, \mmethod~causes relatively more performance degradation to ViT-based detectors compared to CNN-based ones.
This is mainly because of the feature-level slicing strategies, which aim to avoid useless computation on background areas but harm the global modeling by attention blocks in ViT models.
We leave the lossless adaptation to ViT models as our future work.
}

\subsection{Ablation Studies}
\label{sec:exp_abl}

In this subsection, extensive ablation studies are conducted on VisDrone~\cite{zhu2018vision} dataset to validate the superiority of \mmethod's main components, \ie, \moduleA, \moduleB, and \moduleC, on top of the advanced YOLOv5 baseline.

\begin{table}[tb]
    \centering
    \small
    \setlength{\tabcolsep}{5pt}
    \caption{\editf{Ablation on the main components of our method.}}
    \label{tab:abl_slice_sparse}
    \begin{tabular}{ccc|ccc|cc}
        \toprule
        HR & FS & SH & AP$^{s}$ & AP & AP$_{50}$ & GFLOPs & FPS  \\
        \midrule
        - & - & - & 28.5 & 36.2 & 60.1 & 264.9 & 30.5  \\
        \midrule
        $\checkmark$ & - & - & \topa{30.9} & \topa{38.1} & \topa{62.6} & 412.2 & 22.8 \\
        $\checkmark$ & Uni & - & 30.0 & 34.9 & 58.8 & 243.3 & 29.2 \\
        $\checkmark$ & Ada & - & \topa{30.9} & \topa{38.1} & 62.4 & 232.9 & 27.7 \\
        $\checkmark$ & Ada & $\checkmark$ & 30.8 & 37.9 & 62.3 & 180.6 & 28.6 \\
        \midrule
        $\checkmark$ & Sim & - & 30.8 & 37.8 & 62.2 & 236.9 & 29.9 \\
        $\checkmark$ & Sim & $\checkmark$ & 30.7 & 37.7 & 62.0 & \topa{183.5} & \topa{30.9} \\
        \bottomrule
    \end{tabular}
\end{table}

\begin{figure*}[htb]
  \centering
  \includegraphics[width=1.00\linewidth]{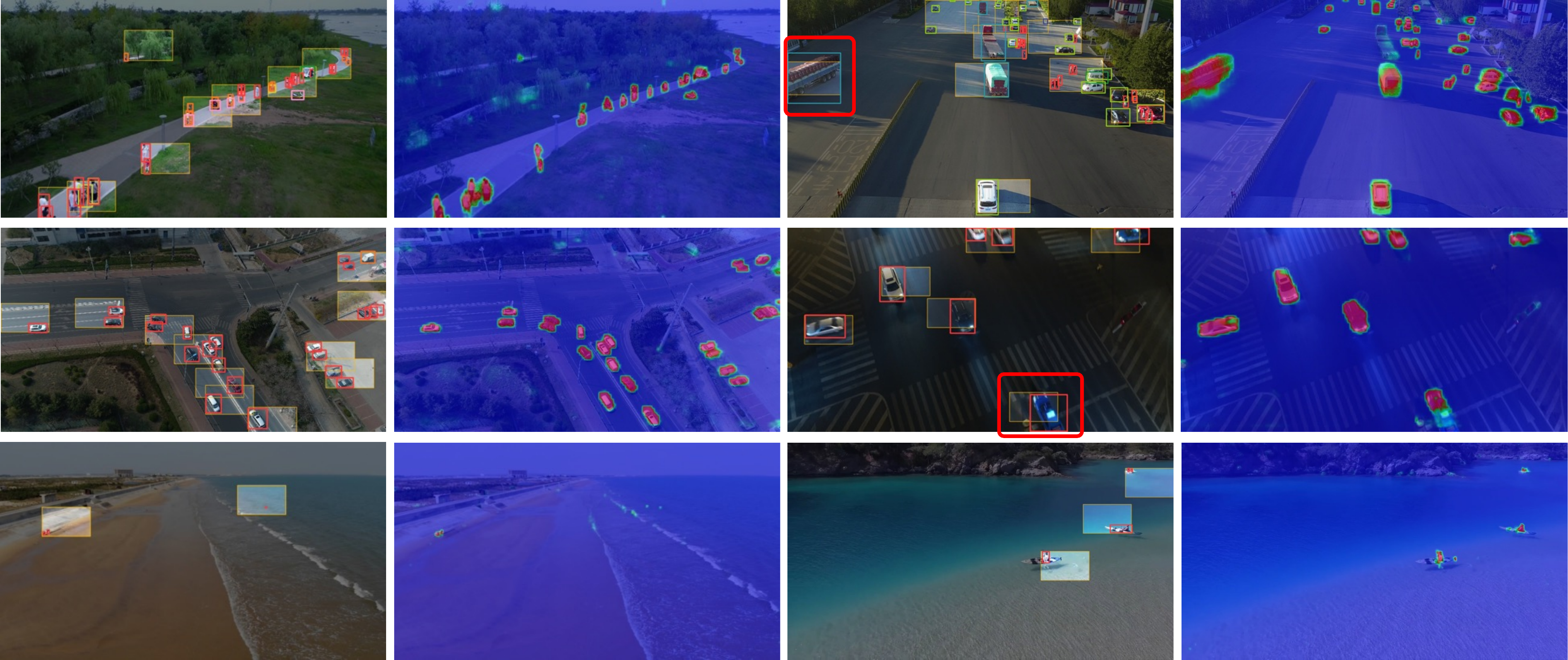}

  \caption{\textbf{Visualization examples on detection results and objectness masks.} Images are selected from VisDrone~\cite{zhu2018vision} (top), UAVDT~\cite{du2018unmanned} (middle), and TinyPerson~\cite{yu2020scale} (bottom), respectively. In the odd columns, object detections are colored by their categories, sliced patches are highlighted within \textcolor[RGB]{255,185,0}{yellow} boxes, and background areas are masked in \textcolor{gray}{gray}. The results illustrate our \mmethod~can effectively and efficiently detect those sparsely clustered small objects. Specifically, the relatively large object exceeding the sliced patch is still completely detected (marked with \textcolor{red}{red} boxes), indicating our method can also handle the scale variation problem.}
  \label{fig:res_vis}
\end{figure*}

\textbf{All of the proposed modules are effective.}
To verify our proposed modules, we first construct the baseline model as \cref{sec:exp_impl} describes. 
Then we simply $1.25\times$ enlarge the input size (from 1,536 to 1,920, denoted as \textit{HigherResolution(HR)}), and the detection performance immediately gains (\eg, 2.4 improvements of AP$^{s}$), as shown in \cref{tab:abl_slice_sparse}.
It is consistent with our motivation that simple resolution-enlarging is an effective way to detect small objects, while the computation cost increases consequently (\eg, 412.2 \vs~264.9 GFLOPs).
Given the objectness mask predicted by \moduleA, uniformly slicing the feature map and discarding the background patches (denoted as \textit{Uni FeatSlicer(FS)}) can reduce around 40\% of computation and speed up the inference from 22.8 FPS to 29.2 FPS.
However, the detection performance dramatically drops (\eg, 3.2 of AP) since numerous objects are truncated by the sliced patches, as discussed in \cref{sec:method_adaslicer}. 
By contrast, our proposed \moduleB~(denoted as \textit{Ada FS}) can reduce the computation with negligible loss of detection precision, and more visualizations are provided in \cref{fig:res_vis}.
The \moduleC~further saves over 50 GFLOPs via sparse convolutions on the detection head.
However, though \moduleB~decrease the computation, the overall inference speed drops due to the unparallelizable operations in \cref{alg:method_slice}.
\editf{
Consequently, we evaluate the simplified \cref{alg:method_slice_sim} as an alternative (denoted as \textit{Sim FS}), and \cref{tab:abl_slice_sparse} shows a considerable acceleration (29.9 \vs~27.7 FPS) powered by GPU parallelization.
However, the average precision for object detection receives a non-negligible decline (\eg, 0.3 decrease of AP), mainly due to the increased truncation in large objects, as discussed in \cref{sec:method_adaslicer}.
Therefore, we suggest adopting \cref{alg:method_slice} upon appropriate engineering optimization, otherwise \cref{alg:method_slice_sim} is an alternative solution for overall inference efficiency.
}

\begin{table}[t]
    \centering
    \small
    \setlength{\tabcolsep}{3pt}
    \caption{Ablation on pseudo-label strategy for training \moduleA.}
    \label{tab:abl_hm_label}
    \begin{tabular}{c|cc|ccc}
        \toprule
        pseudo-label &  BPR$^{\texttt{box}}$ & BPR$^{\texttt{ctr}}$ & AP$^{s}$ & AP & AP$_{50}$ \\
        \midrule
        Gaussian   & 99.1 & \topa{98.3} & 28.2\mdetla{+0.0} & 35.7\mdetla{+0.0} & 59.5\mdetla{+0.0}   \\
        SAM~\cite{kirillov2023segment}     & 98.9 & 97.7 & 27.9\mdetla{-0.3} & 35.9\mdetla[brown]{+0.2} & 59.5\mdetla{+0.0} \\
        \rowhl Hybrid     & \topa{99.3} & \topa{98.3} & \topa{28.3}\mdetla[brown]{+0.1} & \topa{36.0}\mdetla[brown]{+0.3} & \topa{59.7}\mdetla[brown]{+0.2} \\
        \bottomrule
    \end{tabular}
\end{table}

\begin{table}[t]
    \centering
    \small
    \setlength{\tabcolsep}{3pt}
    \caption{Ablation on implementation for the \moduleA~module.}
    \label{tab:abl_hm_module}
    \begin{tabular}{c|cccc|cc}
        \toprule
        Impl. & P$^{\texttt{mask}}$ & R$^{\texttt{mask}}$ & BPR$^{\texttt{box}}$ & BPR$^{\texttt{ctr}}$ & GFLOPs & Latency \\
        \midrule
        Conv    & 89.5 & 82.7 & 98.9 & 97.8 & 12.20 & \topa{0.61}  \\
        DCN~\cite{dai2017deformable}     & 89.5 & \topa{84.6} & \topa{99.3} & \topa{98.4} & 13.94 & 3.54  \\
        SPP~\cite{he2015spatial}     & \topa{90.5} & 83.0 & 99.1 & 98.3 & 3.42 & 1.27  \\
        ASPP~\cite{chen2017deeplab}    & 90.4 & 84.0 & \topa{99.3} & 98.2 & 3.41 & 1.68  \\
        \rowhl DWConv  & 89.8 & 83.3 & \topa{99.3} & 98.3 & \topa{1.22} & 0.76  \\
        \bottomrule
    \end{tabular}
\end{table}

\textbf{Prior knowledge facilitates preliminary object-seeking.}
At the start of our method, the object-seeking is conducted through class-agnostic objectness mask estimation, and hybrid pseudo-masks are generated for training. 
Specifically, we construct Gaussian distributions for pseudo-masks via bounding box annotations, and then use SAM~\cite{kirillov2023segment} predictions to regularize the Gaussian distributions' shapes. 
The goal is to introduce prior knowledge (object's shape) by the generalized SAM model into the learning process.
In fact, as shown in \cref{tab:abl_hm_label}, Gaussian masks are capable enough for preliminary object-seeking (\ie, 99.1\% of BPR$^{\texttt{box}}$ and 98.2\% of BPR$^{\texttt{ctr}}$), while SAM predictions lead to a decline in BPR$^{\texttt{box}}$ and BPR$^{\texttt{ctr}}$ and ultimately the final detection performance on small objects (\ie, 27.9 \vs~28.3 of AP$^s$). 
It is mainly because SAM struggles with segmenting small objects, as discussed in \cref{sec:method_objseeker}.
However, combining Gaussian masks with SAM predictions brings better detection precision (\eg, 0.3\% gains of AP).
We owe it to the integrated shape knowledge (beyond bounding box annotations) from SAM, which facilitates the preliminary feature extraction of the network, resulting in slight improvements in the final performance.

\begin{table}[tb]
    \centering
    \small
    \caption{Ablation on patch-size to slice.}
    \label{tab:abl_slice_size}
    \begin{tabular}{c|ccc|cc}
        \toprule
        $k$ & AP$^{s}$ & AP & AP$_{50}$ & GFLOPs & FPS  \\
        \midrule
        4  & \topa{30.8} & \topa{38.0} & 62.5 & 273.1 & 23.2 \\
        \rowhl 8  & \topa{30.8} & 37.9 & 62.3 & 183.5 & 28.6 \\
        16 & 29.9 & 34.7 & 58.9 & \topa{139.5} & \topa{30.1} \\
        \bottomrule
    \end{tabular}
\end{table}

\begin{figure}[b]
  \centering
  \includegraphics[width=1.00\linewidth]{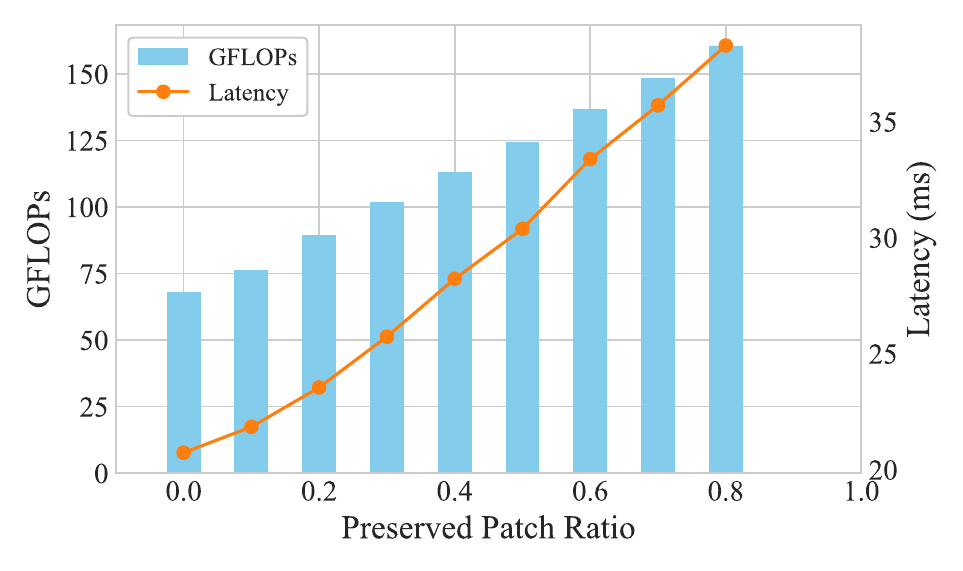}

  \caption{\editf{\textbf{Bucketized statistics of GFLOPs cost and overall latency.}}}
  \label{fig:abl_bucket}
\end{figure}

\textbf{Lightweight module is enough for object-seeking.}
In the \moduleA~module, we simply employ a depth-wise separable convolutional block (DWConv) to predict the objectness mask.
And the primary goal is to seek as many objects as possible. 
According to \cref{tab:abl_hm_module}, the large-kernel DWConv obtains better objectness estimation performance than the standard convolutional block (\eg, about 0.5\% improvement of BPR$^{\texttt{box}}$ and BPR$^{\texttt{ctr}}$) due to the larger receptive field. 
And DWConv achieves comparable best-possible-recalls against SPP~\cite{he2015spatial} and ASPP~\cite{chen2017deeplab} with less computation and latency.
Though DCN~\cite{dai2017deformable} brings higher recall on the predicted objectness mask (\ie, R$^{\texttt{mask}}$), the BPR$^{\texttt{box}}$ and BPR$^{\texttt{ctr}}$ are not considerably increased. However, the latency of 3.54 ms is unacceptable.
Therefore, we claim that the lightweight DWConv block is capable enough for object-seeking.

\textbf{Patch size is a trade-off between efficacy and efficiency.}
According to \cref{tab:abl_slice_size}, if we slice the feature patches at 1/4 size of the original feature map, the computation simultaneously increases by 44\% while the FPS drops from 28.6 to 23.2, as more background areas exist in the sliced patches.
However, the detection performance does not earn a considerable gain (\eg, only a 0.1 increase on AP).
On the contrary, when the patch size becomes 1/16 of the feature map, the computation is reduced while detection precision is also degraded, indicating small patch size leads to more truncation on objects.
Thus, 1/8 is a proper coefficient to decide the patch size on the current VisDrone~\cite{zhu2018vision} dataset.
As for other datasets like TinyPerson~\cite{yu2020scale}, where objects are far more small and sparsely located, 1/16 may be a more suitable choice. 
Overall, the patch size is a trade-off between efficacy and efficiency, which depends on specific datasets.

\editf{
\textbf{Computational costs grow linearly as the preserved patches increase.}
In previous experiments, we merely report the averaged GFLOPs and FPS on each input image in the validation set as the measure proxy.
As the computational cost of our method may vary on different images and datasets (depending on the object size and density in images), we further employ a bucketization strategy to measure the GFLOPs and latency on images with the preserved patch ratio (by our \moduleA) ranging from 0\% to 100\%.
As illustrated in \cref{fig:abl_bucket}, the costs grow nearly linearly as the preserved patches increase, which is consistent with our empirical results that our method saves more computation on TinyPerson~\cite{yu2020scale} than VisDrone~\cite{zhu2018vision} datasets, as the former has fewer foreground objects/patches.
One can predict how our \mmethod~can benefit small object detection according to the object distribution in target application scenarios.
}

\begin{figure*}[htb]
  \centering
  \includegraphics[width=1.00\linewidth]{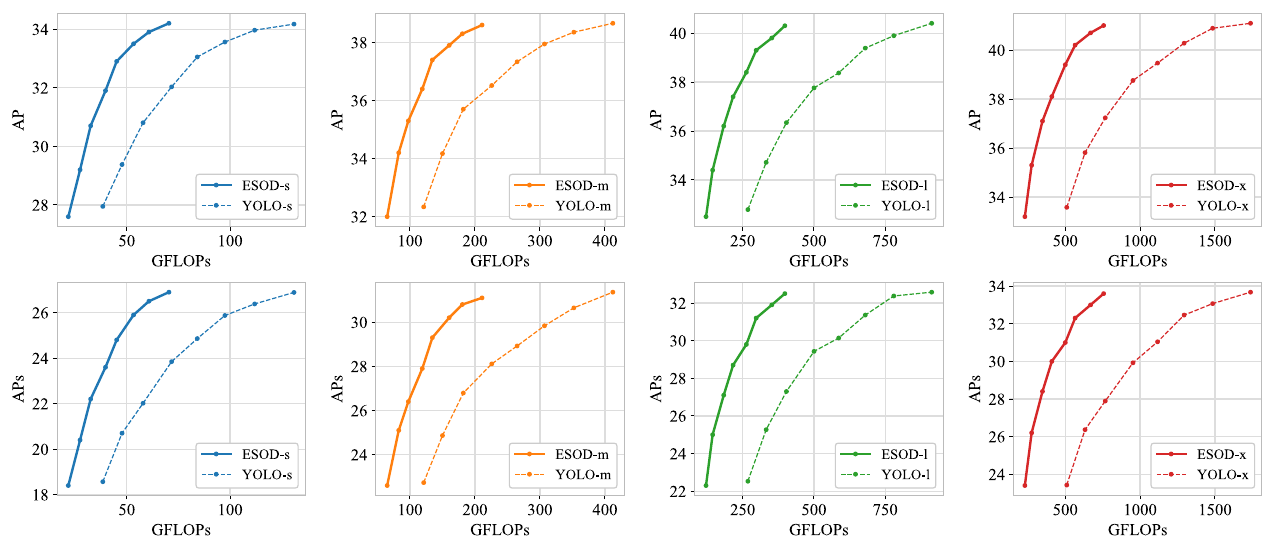}

  \caption{\textbf{Performance comparison with different size of backbones. } Our \mmethod~persistently surpasses the baseline detector by a large margin.}
  \label{fig:abl_yolo_all}
\end{figure*}

\textbf{Our method effectively scales up to larger networks.}
Our \mmethod~is mainly built with the medium-size backbone~\cite{glenn_jocher_2021_4679653} for better performance, but it does not mean \mmethod~is not suitable for other network sizes.
Actually, \cref{fig:abl_yolo_all} illustrates our method persistently outperforms the baseline with all of the backbone series (\ie, \textbf{s}mall-, \textbf{m}edium-, \textbf{l}arge-, and \textbf{x}large-sizes) by a large margin. 
It implies our \mmethod~can be implemented with different sizes of networks according to the computation resources.
For example, one may employ the small-size backbone on edge devices while adopts the large-size backbone on GPU servers.

\subsection{Visualizations}
\label{sec:exp_vis}

To qualitatively demonstrate the efficacy of our \mmethod, some representative examples are displayed in \cref{fig:res_vis}, including predicted objectness masks, sliced patches, and final detection results.
It shows that the sparsely clustered small objects are coarsely but effectively recognized, and the massive background regions are discarded.
In the sliced patches, small objects are successfully detected.

It is worth noting that even though some large objects are visually truncated by patches, the final predictions are still complete (the top-right example). 
\editf{In fact, as discussed in \cref{sec:method_adaslicer}, network's receptive field can exceed the feature patches after the preliminary feature extraction process, and our \moduleB~determines the patches centered at large objects' centers. 
In this way, a large proportion of those objects are enclosed within the sliced patches, making the detection complete.}
Therefore, our \mmethod~can concurrently accelerate small object detection while keep relatively large objects detected.


\section{Conclusion}
\label{sec:conclu}

In this paper, we statistically point out that in practice, numerous small objects are sparsely distributed and locally clustered in high-resolution images. 
Rather than subtle feature finetuning, image enlargement is more effective. We are committed to saving computations and time costs, so as to enlarge input images for small object detection.
Specifically, we conduct the preliminary object-seeking and adaptive patch-slicing at the feature level via \moduleA~and \moduleB, where redundant feature extraction is avoided.
Incorporating sparse convolutions, \moduleC~reuses the predicted objectness mask for sparse detection.
The resulting method, namely \mmethod, is able to reduce the massive computation and GPU memory wasted on feature extractions and object detection on background areas. 
In addition, our \mmethod~is a generic framework for both CNN- and ViT-based networks.
Experiments on VisDrone, UAVDT, and TinyPerson datasets illustrate that our \mmethod~vastly reduces the computation costs and significantly outperforms the state-of-the-art competitors.

\section*{Acknowledgment}

This work was supported in part by the Fundamental Research Funds for the Central Universities, in part by Alibaba Cloud through the Research Intern Program, and in part by Zhejiang Provincial Natural Science Foundation of China under Grant No. LDT23F01013F01.
The work of Kai Liu was completed when he was with Alibaba Cloud.

{\small
\bibliographystyle{IEEEtran}
\bibliography{ref}
}

\vfill

\end{document}